\documentclass[conference]{IEEEtran}
\IEEEoverridecommandlockouts
\usepackage{cite}
\usepackage{amsmath,amssymb,amsfonts}
\usepackage{algorithmic}
\usepackage{graphicx}
\usepackage{textcomp}
\usepackage{xcolor}
\usepackage{booktabs}
\usepackage{multirow}
\def\BibTeX{{\rm B\kern-.05em{\sc i\kern-.025em b}\kern-.08em
    T\kern-.1667em\lower.7ex\hbox{E}\kern-.125emX}}

\usepackage{hyperref}

\newcommand{\dq}[1]{``#1''}

\begin{document}

\title{Comparative Analysis of Liquid Neural Networks and LSTM for Sequential Pattern Recognition: Robustness, Efficiency, and Clinical Utility\\
{\footnotesize \textsuperscript{*}Extended preprint version. The conference paper will appear in Proceedings of JCSSE 2026 (June 24--27, 2026, Bangkok, Thailand).}
}

\author{
\IEEEauthorblockN{
Ye Kyaw Thu$^{1,2}$\thanks{Corresponding author: yekyaw.thu@nectec.or.th},
Thazin Myint Oo$^{2}$,
Thepchai Supnithi$^{1}$
}

\IEEEauthorblockA{
$^{1}$National Electronics and Computer Technology Center (NECTEC), Pathumthani, Thailand
}

\IEEEauthorblockA{
$^{2}$Language Understanding Lab., Yangon, Myanmar
}

\IEEEauthorblockA{
Email: yekyaw.thu@nectec.or.th, queenofthazin@gmail.com, thepchai.supnithi@nectec.or.th
}
}

\maketitle

\begin{abstract}
Traditional Recurrent Neural Networks (RNNs) and Long Short-Term Memory (LSTM) units operate on discrete time steps, often failing to capture the fluid temporal dynamics of real-world physical processes. Liquid Neural Networks (LNNs), specifically Closed-form Continuous-time (CfC) networks, address this by modeling the hidden state evolution as a continuous differential equation. In this paper, we conduct a comprehensive benchmarking study across four distinct sequential modalities: neuromorphic event-based data (N-MNIST), stroke-based drawing (QuickDraw), visual handwriting (IAM), and physiological time-series (PhysioNet Sepsis-3). Furthermore, we perform a rigorous stress test using temporal dropout to evaluate model robustness against missing data. Our findings reveal that LNNs consistently provide superior parameter efficiency and significantly higher robustness in natively temporal domains and clinical environments where data sparsity is prevalent. This extended preprint provides additional background on related datasets and the LNN theoretical lineage, supplemented with a detailed appendix documenting our full implementation and experimental settings.
\end{abstract}

\begin{IEEEkeywords}
Liquid Neural Networks, CfC, LSTM, Sepsis Prediction, Robustness, Temporal Dropout, Neuromorphic Computing, Sequential Pattern Recognition.
\end{IEEEkeywords}

\section{Introduction}

Deep learning for sequential data has long been dominated by the Long Short-Term Memory (LSTM) network \cite{lstm1997}. While modern variants like the GRU \cite{gru2014} have optimized these architectures, they remain rooted in the fundamental principles of backpropagation through time established decades ago \cite{rumelhart1986}. However, LSTMs operate as discrete-time operators, treating time simply as an index in a sequence. This \dq{grid-based} view of time is inherently problematic for data captured from continuous physical systems, such as high-frequency physiological vital signs, variable-rate stroke data, or event-based neuromorphic cameras.

Liquid Neural Networks (LNNs) represent a paradigm shift inspired by the remarkably efficient nervous system of the C. elegans roundworm. By utilizing Neural Circuit Policies (NCPs) \cite{lechner2020ncp}, researchers have demonstrated that sparse, biologically-inspired wiring can achieve complex navigation behaviors that current robotic systems struggle to replicate with millions of parameters. LNNs emulate this by using a \dq{liquid} hidden state governed by an Ordinary Differential Equation (ODE) framework, where the network's time-constant is not a fixed hyperparameter but a dynamic function of the input.

In this work, we demonstrate that this continuous-time formulation provides a critical advantage in \textit{robustness} and \textit{temporal stability}. We conduct a comprehensive benchmarking study across four diverse sequential modalities. Beyond standard performance metrics, we introduce a \dq{temporal stress test} to evaluate how these models perform when faced with real-world sensor failures or irregular sampling rates. Our findings suggest that LNNs are better suited for mission-critical applications---such as clinical monitoring and autonomous navigation---where data integrity is not guaranteed.

\section{Related Work}
\subsection{From Discrete to Continuous-Time RNNs}

The evolution of sequential modeling moved from standard RNNs to gated architectures like LSTMs \cite{lstm1997} and GRUs \cite{gru2014} to mitigate vanishing gradients. However, these remain discrete. The field of Neural Ordinary Differential Equations (Neural ODEs) \cite{chen2018neuralode} introduced the idea of modeling the hidden state transition as a continuous flow. While powerful, Neural ODEs often require expensive numerical solvers that increase training latency.

\subsection{Liquid Neural Networks and CfC}

LNNs emerged as a sub-class of continuous-time models, specifically the Liquid Time-constant (LTC) networks \cite{hasani2021ltc}. Unlike standard Neural ODEs, LTCs possess a varying time-constant that adapts to the input's temporal resolution, building upon the theoretical framework of interpretable continuous-time control \cite{hasani2020phd}. The Closed-form Continuous-time (CfC) model \cite{hasani2022cfc} used in this paper represents the latest advancement, replacing numerical ODE integration with a closed-form analytical solution.

\subsection{Applications in Robust and Dynamic Environments}

Recent studies have showcased LNNs in high-stakes environments. For instance, LNNs have demonstrated superior \dq{out-of-distribution} generalization in drone flight navigation \cite{chahine2023flight}. Furthermore, emerging research explores their utility in telecommunications for optimizing next-generation wireless networks \cite{zhu2025telecom}.


\subsection{Biological Foundations: C.~elegans and Neural Circuit Policies}

A key conceptual pillar of LNNs is the nervous system of \textit{Caenorhabditis elegans} (C.~elegans), a nematode whose complete connectome of 302 neurons and roughly 7,000 synaptic connections has been fully mapped \cite{white1986celegans}. Despite this extreme simplicity, the organism exhibits robust locomotion, chemotaxis, and thermotaxis behaviors. This parsimony motivated Lechner et al.\ \cite{lechner2020ncp} to propose Neural Circuit Policies (NCPs): sparsely-wired recurrent networks whose inter-neuron connectivity is drawn from biologically-plausible patterns. NCPs demonstrated that sparse, structured wiring enables a robot to navigate complex visual environments with far fewer parameters than conventional deep networks. LNNs extend this insight by coupling the NCP wiring philosophy with a continuous-time ODE dynamics, yielding models that are simultaneously interpretable, robust, and parameter-efficient.

\subsection{Liquid Time-constant Networks: From LTC to CfC}

The formal theoretical lineage leading to the CfC model used in this paper spans three generations of continuous-time RNNs. The original Liquid Time-constant (LTC) network \cite{hasani2021ltc} defined each neuron's hidden state $h(t)$ via a first-order ODE whose \textit{time constant} $\tau$ is itself a function of the current input $x(t)$:
\begin{equation}
\tau(x,t)\,\frac{dh(t)}{dt} = -h(t) + f\!\left(x(t),\, h(t),\, W,\, b\right),
\end{equation}
where $\tau(x,t)>0$ ensures numerical stability. This input-dependent time constant makes LTCs \dq{liquid}: the network intrinsically slows down or speeds up its internal dynamics according to the informational density of the incoming signal. However, evaluating the ODE requires a numerical integrator (e.g., Euler or Runge-Kutta), which introduces high per-step computational cost during training.

To overcome this bottleneck, Hasani et al.\ \cite{hasani2022cfc} derived a closed-form solution that replaces the ODE solver. The CfC approximation expresses the hidden state at any future time $t+\Delta t$ directly as a nonlinear function of $h(t)$, $x(t)$, and $\Delta t$, without numerical integration. This brings the computational cost of continuous-time modeling on par with standard LSTMs while preserving the liquid temporal-adaptation property. Hasani's earlier dissertation \cite{hasani2020phd} provides extensive theoretical groundwork connecting Lyapunov stability theory to these continuous-time recurrent designs.

A parallel line of research by Lechner and Hasani \cite{lechner2020learning} demonstrated that this continuous-time formulation facilitates learning long-term dependencies in open-ended numerical expressions, a task where discrete-time models typically fail due to gradient attenuation over many steps.

\subsection{Neuromorphic Event-based Vision and the N-MNIST Dataset}

Neuromorphic vision sensors---also called Dynamic Vision Sensors (DVS) or event cameras---operate on an entirely different principle from frame-based cameras. Rather than capturing pixel intensities at a fixed frame rate, a DVS independently fires an asynchronous \dq{event} $(x, y, t, p)$ for each pixel whenever its log-intensity changes beyond a threshold, where $p \in \{+1,-1\}$ encodes the polarity of the change \cite{lichtsteiner2008dvs}. This produces sparse, high-temporal-resolution streams that encode \textit{motion and temporal contrast} rather than absolute brightness, making the data inherently sequential and asynchronous.

The N-MNIST dataset \cite{nmnist2015} was constructed by recording the standard MNIST digit images with a DVS while the sensor executes three pre-defined saccadic eye movements. The resulting event streams share the same 10-class label space as MNIST but carry spatiotemporal information absent in static images. N-MNIST has since become a standard neuromorphic benchmark, with published results ranging from simple integrate-and-fire SNNs to convolutional event-frame approaches \cite{sironi2018hats}. A recurring challenge is that frame-based feature extractors must first bin the asynchronous events into temporal slices before applying spatial convolutions, inevitably discarding sub-bin temporal resolution. In our implementation, we accumulate events into 10 temporal bins per sample, applying a log-compression $\log(1 + \text{count})$ to reduce the dynamic range of high-activity regions, before feeding the resulting $10 \times 2 \times 34 \times 34$ tensor to the CNN-RNN backbone.

\subsection{Stroke-based Sketch Recognition and the Google QuickDraw Dataset}

Human sketch recognition has a long history in the pattern recognition community, from early work on online handwriting to modern deep sequence models \cite{graves2013handwriting}. The core challenge in stroke-based recognition is that drawings are inherently \textit{sequential}: the category label depends not just on which pixels are filled, but on the temporal order and spatial trajectory of pen strokes.

The Google QuickDraw dataset \cite{quickdraw2017}, collected from the \textit{Quick, Draw!} web game, contains over 50 million crowd-sourced sketches across 345 categories, each stored as a variable-length sequence of stroke vectors. Ha and Eck \cite{quickdraw2017} introduced a sequence-to-sequence generative model (Sketch-RNN) that learns to \textit{generate} new sketches, demonstrating the viability of recurrent models for vector-graphic sketch data. Subsequent work has applied attention mechanisms and graph-based representations \cite{xu2021sketchbert} to achieve strong classification performance. Our work takes a simpler approach, treating the stroke sequence as a direct RNN input and comparing the classification accuracy of LNN (CfC) and LSTM cores on a 10-class subset, specifically to isolate the temporal modeling capability rather than architectural complexity.

\subsection{IAM Handwriting Database and Offline Recognition}

The IAM Handwriting Database \cite{iam2002} is one of the oldest and most widely used benchmarks in offline handwriting recognition. It contains handwritten forms of Lancaster-Oslo/Bergen (LOB) corpus sentences written by 657 different writers, with line-level, word-level, and character-level segmentations. A standard evaluation metric for sequence labeling tasks such as handwriting recognition is the Character Error Rate (CER), defined as the edit distance between the predicted and reference character sequences, normalized by the reference length.

State-of-the-art systems on IAM combine convolutional feature extractors with bidirectional LSTM cores and CTC decoding \cite{graves2009novel}. More recent approaches employ Transformer-based architectures \cite{li2021trocr}, achieving CERs below 4\% at the line level. However, these high-performing models require much larger training sets and bidirectional processing. In our study, we deliberately use a compact ResNet-6 backbone with a unidirectional sequential core (either CfC or LSTM) to expose any differences attributable to the temporal modeling mechanism alone. The CTC loss \cite{ctc2006} is used throughout to handle the alignment between the visual feature sequence and the unsegmented character string.

\subsection{PhysioNet/CinC Challenge 2019 and Sepsis-3 Clinical Context}

The PhysioNet/Computing in Cardiology Challenge 2019 \cite{physionet2019} released de-identified ICU patient records from two hospital systems, containing up to 40 time-varying clinical variables (vital signs and lab results) sampled at irregular intervals. The task is early prediction of sepsis as defined by the Sepsis-3 consensus \cite{singer2016sepsis}, which requires the presence of life-threatening organ dysfunction caused by a dysregulated host response to infection.

Sepsis prediction from physiological time-series is a well-studied but clinically challenging problem due to three key characteristics: (1) \textit{class imbalance}---sepsis patients are a minority in most ICU cohorts; (2) \textit{irregular sampling}---clinical measurements are taken on a need-driven basis rather than at fixed rates; and (3) \textit{missing values}---it is common for many physiological variables to have no recorded value for large stretches of a patient's stay. Prior work has applied standard LSTMs \cite{harutyunyan2019multitask}, Transformer-based models \cite{zhang2020sepsis}, and GRU-based event networks \cite{rubanova2019latent} to this domain. Our work adds a new data point: a controlled head-to-head comparison between LNN (CfC) and LSTM on the same feature set under identical training conditions, specifically examining the false-positive rate as a clinical deployment criterion.


\section{Data Modalities and Benchmark Environments}
To evaluate the flexibility of Liquid Neural Networks (LNNs) against LSTMs, we selected four datasets representing distinct temporal modalities: asynchronous event streams, stroke-based coordinates, visual-to-sequence images, and irregular clinical time-series.

\subsection{Neuromorphic Event Data (N-MNIST)}
Unlike traditional frame-based vision data, the N-MNIST dataset \cite{nmnist2015} consists of asynchronous events captured by a Dynamic Vision Sensor (DVS).
Each sample is represented as a stream of events $(x, y, t, p)$, where $(x, y)$ denote pixel location, $t$ is the timestamp, and $p$ indicates event polarity.
Events are generated as the original MNIST image is moved across the sensor using a predefined saccadic motion.

\begin{figure}[htbp]
    \centering
    \includegraphics[width=0.95\linewidth]{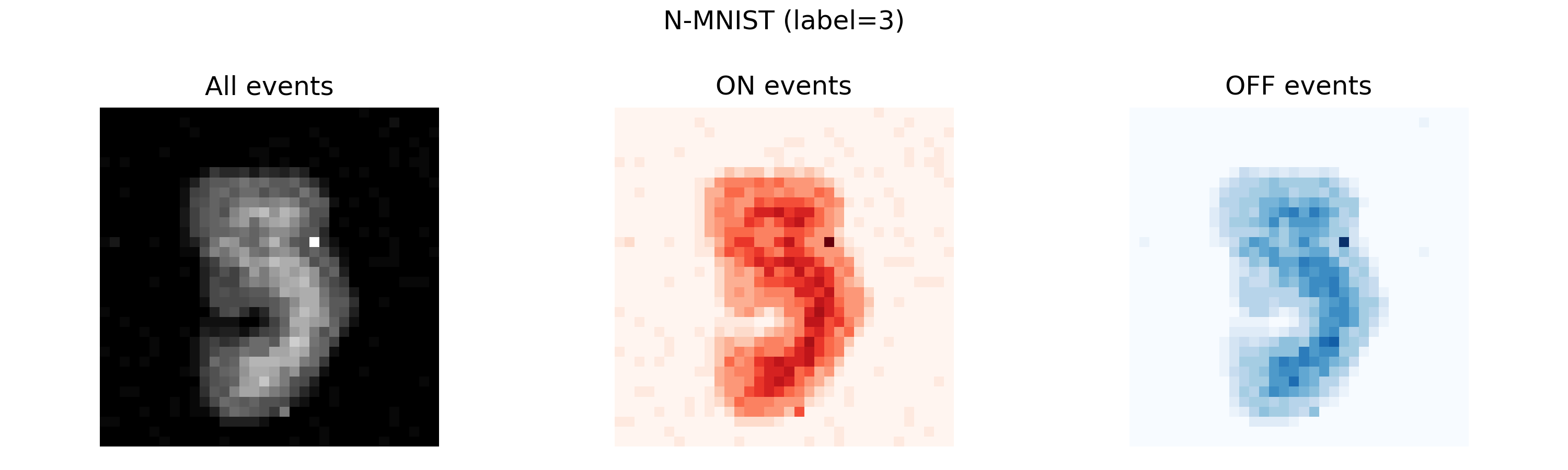}\\
    \vspace{0.3em}
    \includegraphics[width=0.95\linewidth]{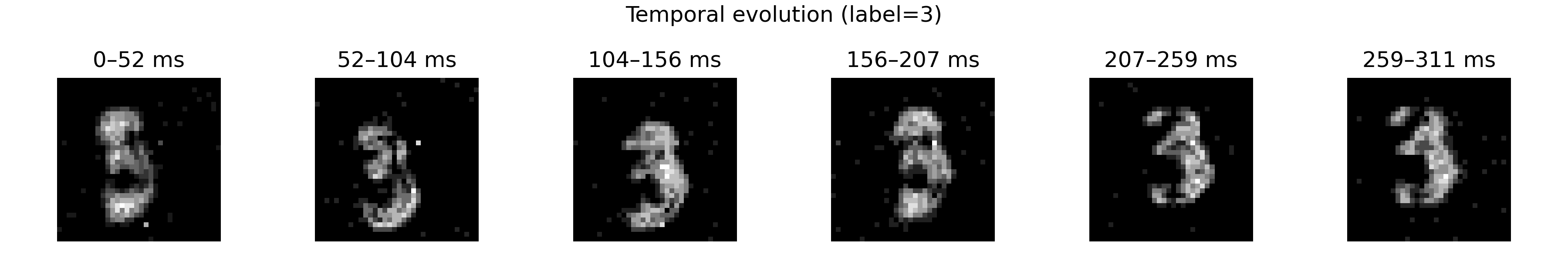}
    \caption{
    Example visualization of an N-MNIST sample (digit ``3'').
    \textbf{Top:} Accumulated event image over the full recording duration, showing all events (left), ON events (middle), and OFF events (right).
    \textbf{Bottom:} Temporal slices obtained by accumulating events over successive time windows, illustrating the spatio-temporal evolution caused by saccadic motion.
    }
    \label{fig:nmnist_data}
\end{figure}

Figure~\ref{fig:nmnist_data} illustrates an example N-MNIST sample.
By accumulating events over time, a spatial representation resembling the original digit emerges, while separating ON and OFF events highlights polarity-specific activity.
Temporal slices further demonstrate the inherently spatio-temporal nature of the data, where meaningful information is encoded in both the timing and location of events.
This dataset therefore evaluates a model's ability to process sparse, high-temporal-resolution neuromorphic signals.

\subsection{Stroke-Based Drawings (Google QuickDraw)}
The Google QuickDraw dataset represents drawings as sequences of pen movements rather than static images.
Each sample consists of a variable-length sequence of stroke points encoded as $(\Delta x, \Delta y, p)$, where $(\Delta x, \Delta y)$ denotes the relative pen displacement and $p$ indicates the pen state (pen-down or pen-up).
This representation emphasizes the temporal ordering of strokes and requires models to integrate local movements over time to recover global shape information.

In this work, we use a subset of the QuickDraw dataset consisting of ten object categories (e.g., airplane, apple, banana, and butterfly), following a controlled experimental setup.

\begin{figure}[htbp]
    \centering
    \includegraphics[width=0.95\linewidth]{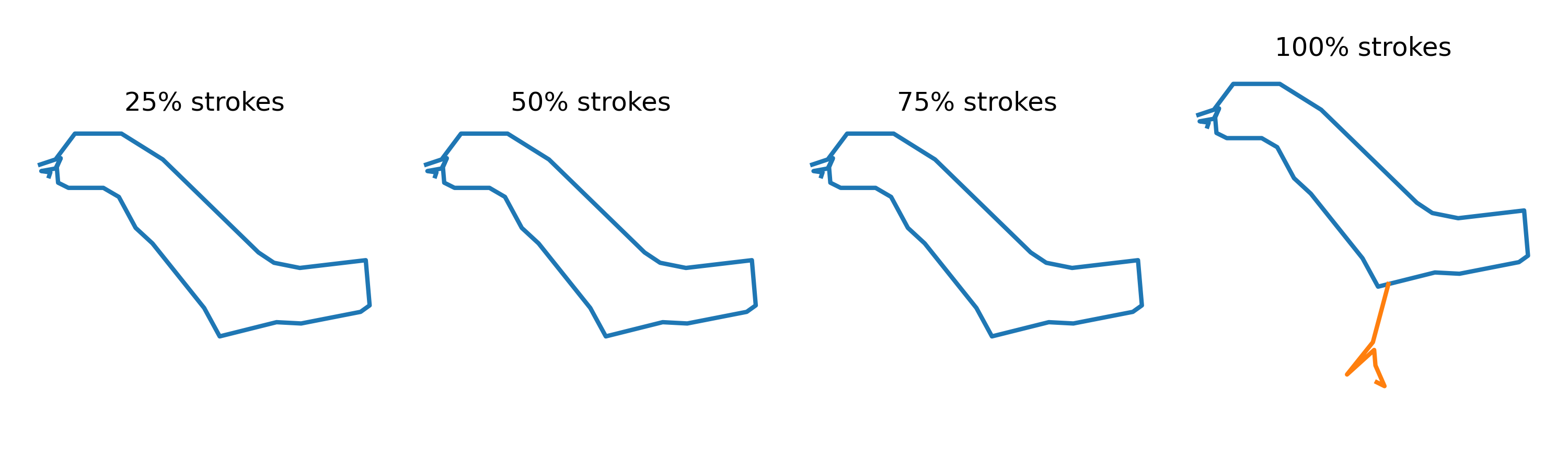}\\
    \vspace{0.2em}
    \includegraphics[width=0.95\linewidth]{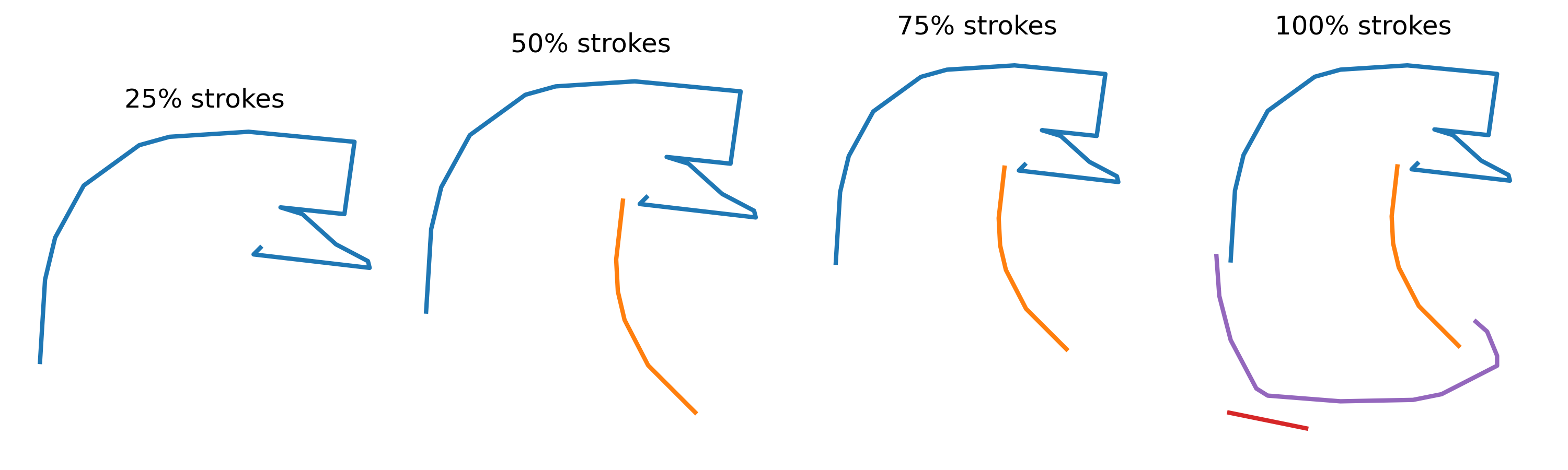}\\
    \vspace{0.2em}
    \includegraphics[width=0.95\linewidth]{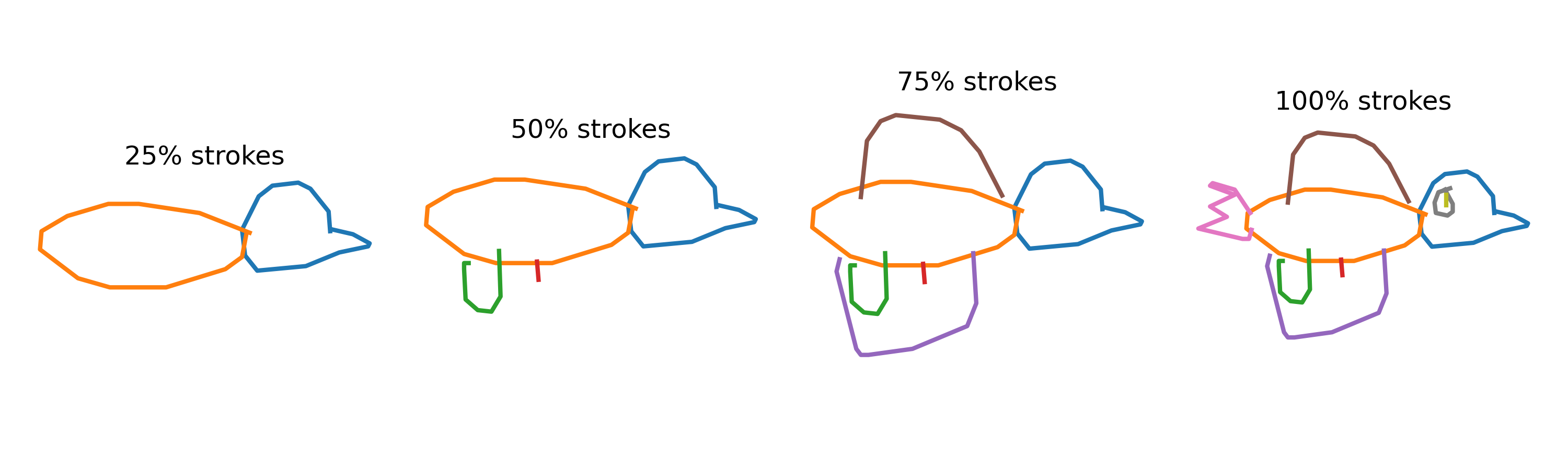}
    \caption{
    Sequential reconstruction of three ``bird'' sketches from the QuickDraw dataset.
    Each row shows the progressive accumulation of stroke points from left (early strokes) to right (complete drawing), illustrating the high intra-class variability in stroke order and spatial structure.
    }
    \label{fig:quickdraw_data}
\end{figure}

Figure~\ref{fig:quickdraw_data} illustrates the sequential nature of QuickDraw data through multiple step-by-step reconstructions of drawings from the same category (``bird'').
Each example is progressively formed by accumulating stroke segments over time, where early time steps contain only partial information and later steps reveal the complete object.
Despite belonging to the same class, the drawings exhibit noticeable variability in stroke order and structure, highlighting the challenge of modeling variable-length sequences with long-term structural dependencies.

\subsection{Visual Handwriting (IAM)}
The IAM dataset \cite{iam2002} provides line-level images of handwritten English text.
Each image is a grayscale rendering of a full line sentence written by one of 657 writers, exhibiting significant variation in writing style, ink density, and character spacing.

\begin{figure}[htbp]
    \centering
    \includegraphics[width=0.95\linewidth]{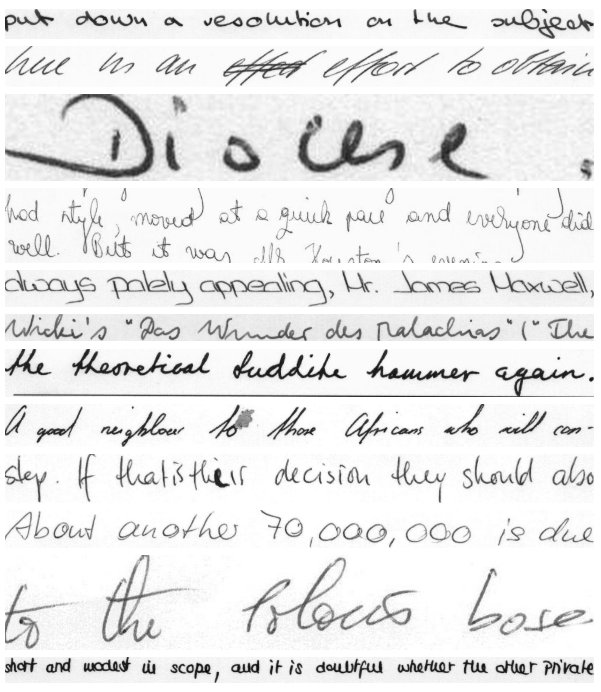}
    \caption{
    Example line images from the IAM Handwriting Database, showing variation in writing style, character spacing, and ink density across different writers.
    }
    \label{fig:iam_data}
\end{figure}

As illustrated in Fig.~\ref{fig:iam_data}, the dataset captures diverse writing styles across different individuals.
The recognition task requires decoding the character sequence from a continuous visual stream without explicit character-level segmentation, which is handled using a Connectionist Temporal Classification (CTC) decoder \cite{ctc2006}.
This modality evaluates a model's ability to process spatial features sequentially over a longer temporal horizon.

\subsection{Physiological Time-Series (PhysioNet Sepsis-3)}
The PhysioNet/CinC Challenge 2019 dataset \cite{physionet2019} contains hourly ICU records with 40 clinical variables per patient. The binary classification target is early sepsis onset as defined by the Sepsis-3 consensus \cite{singer2016sepsis}.

\begin{figure}[htbp]
    \centering
    \includegraphics[width=0.95\linewidth]{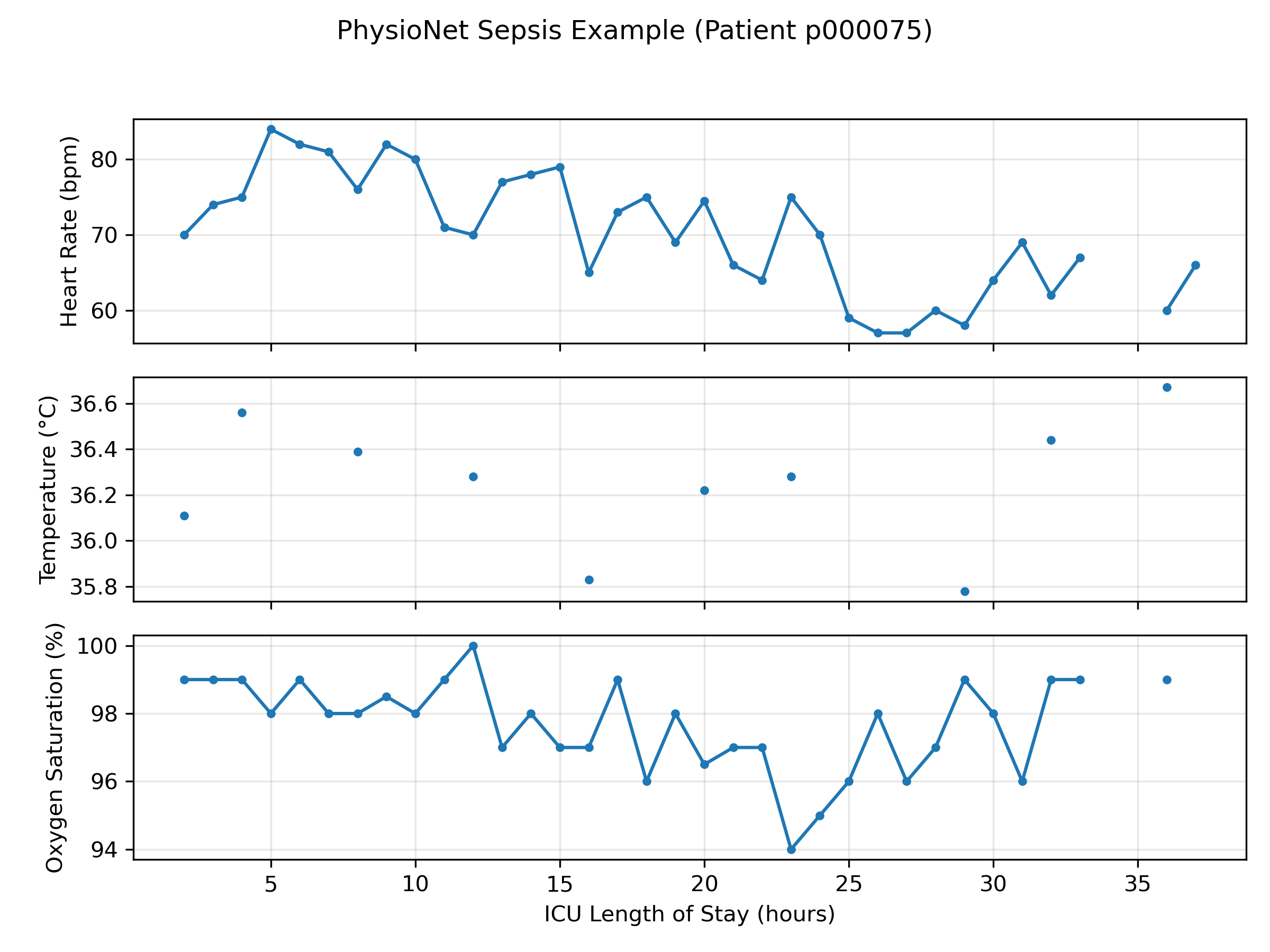}
    \caption{
    Example physiological time-series from the PhysioNet Sepsis-3 dataset.
    Each line represents one clinical variable (e.g., heart rate, temperature) over time.
    Measurements are irregularly sampled and contain missing values.
    The vertical dashed line indicates the onset of sepsis according to the Sepsis-3 definition.
    }
    \label{fig:physio_data}
\end{figure}

As shown in Fig.~\ref{fig:physio_data}, the physiological variables exhibit irregular sampling and frequent missing values.
The dataset is inherently imbalanced, with sepsis-positive patients representing a small fraction of the total patient population.
The time-delta between consecutive observations is included as an additional input feature to make the model aware of sampling irregularity---a critical design choice that allows continuous-time models to exploit the gap duration directly.

\section{Theoretical Framework}

\subsection{From Discrete Gates to Continuous States}
Traditional Recurrent Neural Networks (RNNs) and Long Short-Term Memory (LSTM) units operate on discrete time steps, which often struggle to capture the fluid temporal dynamics of high-frequency physical processes. In contrast, Liquid Neural Networks (LNNs) model hidden state evolution as a continuous-time dynamical system governed by first-order ordinary differential equations (ODEs). Unlike LSTMs, which apply a fixed discrete update per time step, CfC computes an analytical state transition conditioned on elapsed time. As noted in \cite{lechner2020learning}, this approach allows the model to learn long-term dependencies within open-ended numerical expressions more effectively than discrete models. 

The hidden state $h(t)$ in a liquid neuron is defined as:
\begin{equation}
\frac{dh(t)}{dt} = -[A + f(x(t), \theta)] \odot h(t) + f(x(t), \theta) \odot L
\end{equation}
where $f(x(t), \theta)$ modulates the system's time constant, $A$ is the leakage rate, and $L$ is the target state. To maintain computational efficiency, we employ the Closed-form Continuous-time (CfC) solution \cite{hasani2022cfc}. This architecture utilizes non-linear gates---often regularized with dropout \cite{dropout2014} to ensure robustness---to approximate the ODE solution without requiring numerical solvers.

\subsection{The Liquid Property}
The term \dq{Liquid} refers to the model's ability to adjust its internal dynamics based on the input $x(t)$. In a standard RNN, the impact of $x(t)$ on the hidden state is additive. In an LNN, the input influences the \textit{speed} at which the hidden state evolves. This is modeled by an input-dependent time constant $\tau(x, t)$, which ensures that the model can handle irregularly sampled data by naturally interpolating between time steps. This property is what allows LNNs to remain stable during our \dq{temporal dropout} stress tests, as the model's ODE-based foundation treats missing data as a gap in a continuous flow rather than a missing index in a discrete array.

\section{System Architectures}
We designed task-specific backbones for each experiment to ensure a fair comparison between the LSTM and LNN (CfC) cores.

\subsection{Neuromorphic Vision (N-MNIST)}
\begin{figure}[htbp]
    \centering
    \includegraphics[width=0.5\linewidth]{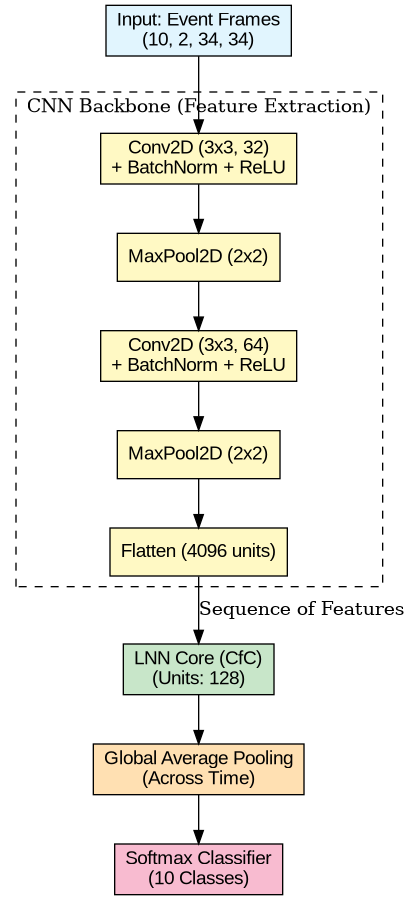}
    \caption{N-MNIST Architecture: CNN backbone with 128 units CfC/LSTM core.}
    \label{fig:arch_nmnist}
\end{figure}

We utilize the N-MNIST dataset \cite{nmnist2015}, which converts static digits into event-based spikes. The architecture (Fig. \ref{fig:arch_nmnist}) utilizes a CNN backbone \cite{lecun1998cnn} to extract spatial features from event-based frames. The raw spikes are accumulated into 10 temporal bins of $34 \times 34$ frames with 2 polarity channels. A two-layer convolutional network (Conv2D with 32/64 filters, BatchNorm, and ReLU) processes these frames, culminating in a flattened 4096-dimensional feature vector per time step. This sequence is then fed into a sequential core of 128 units (either CfC or LSTM). Finally, Global Average Pooling is applied across the temporal dimension to aggregate the latent representations before a 10-way Softmax classifier determines the digit.

\subsection{Handwriting Recognition (IAM)}
\begin{figure}[htbp]
    \centering
    \includegraphics[width=0.30\linewidth]{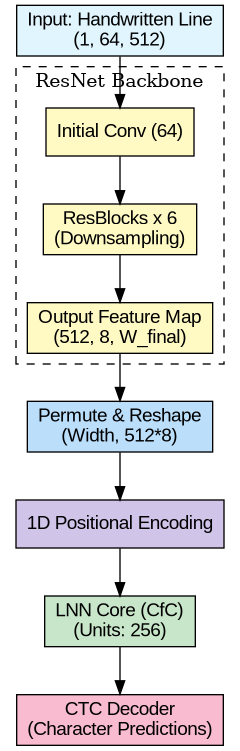}
    \caption{IAM Architecture: ResNet-6 backbone with 256-unit sequential core.}
    \label{fig:arch_iam}
\end{figure}

The IAM pipeline \cite{iam2002} uses a ResNet-6 feature extractor to convert handwritten line images (input size $64 \times 512$) into sequences of visual descriptors (Fig. \ref{fig:arch_iam}). The backbone consists of an initial 64-filter convolution followed by six residual blocks that downsample the spatial dimensions to a feature map of size $512 \times 8 \times W_{final}$. We permute and reshape this map into a 1D sequence of visual features. To preserve sequence order, 1D Positional Encoding is added before the 256-unit CfC or LSTM core. The final predictions are generated via a CTC (Connectionist Temporal Classification) decoder, which handles the alignment between the visual sequence and the character string.

\subsection{Stroke-based Recognition (QuickDraw)}
\begin{figure}[htbp]
    \centering
    \includegraphics[width=0.3\linewidth]{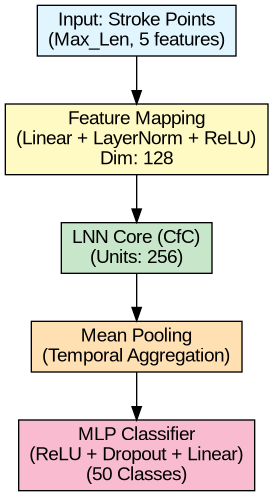}
    \caption{QuickDraw Architecture: 256-unit core for 50-class sketch recognition.}
    \label{fig:arch_quickdraw}
\end{figure}

For the QuickDraw \cite{quickdraw2017} experiment, as illustrated in Fig. \ref{fig:arch_quickdraw}, the model processes raw stroke data consisting of $x, y$ coordinates and pen states. The input features (dim: 5) are first passed through a linear mapping layer combined with LayerNorm and ReLU activation to project them into a 128-dimensional space. This enriched representation is processed by a 256-unit sequential core that models the velocity and trajectory curvature of the drawing. After temporal aggregation via Mean Pooling, a multi-layer perceptron (MLP) classifier featuring Dropout and ReLU layers performs the final 10-class recognition.

\subsection{Physiological Monitoring (Sepsis-3)}

\begin{figure}[htbp]
    \centering
    \includegraphics[width=0.5\linewidth]{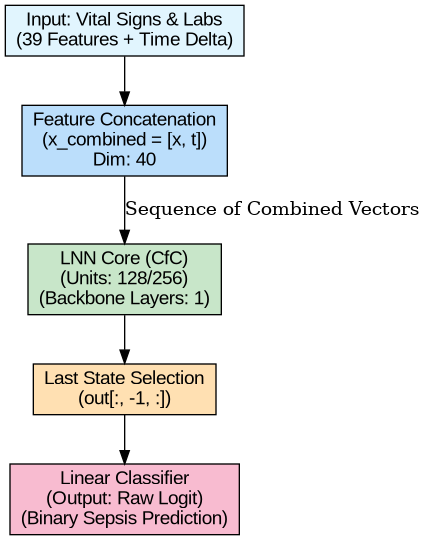}
    \caption{PhysioNet Architecture: 128/256-unit CfC core for binary sepsis prediction.}
    \label{fig:arch_physio}
\end{figure}

The PhysioNet architecture \cite{physionet2019}, as depicted in Fig. \ref{fig:arch_physio}, processes 40 input variables, comprising 39 physiological vital signs and laboratory measurements concatenated with a time-delta feature  \cite{singer2016sepsis}. The core sequential model utilizes either 128 or 256 CfC/LSTM units to track the continuous evolution of a patient's clinical state. Unlike the pooling methods used in vision tasks, this model utilizes the last hidden state of the sequence to feed a linear classifier. This design choice focuses on the most recent clinical indicators to produce a binary logit for sepsis prediction.

\section{Experimental Results}

\subsection{Performance Metrics}
Table \ref{tab:comparison} summarizes the final metrics after training across the N-MNIST, QuickDraw, and IAM datasets. On the natively temporal N-MNIST dataset, the LNN (99.38\% test accuracy) outperforms the LSTM (99.13\%), demonstrating its superior ability to handle event-based visual data where time is a primary feature. While the LSTM maintains a slight lead in accuracy for the QuickDraw and IAM datasets, the LNN provides a highly competitive and more parameter-efficient alternative. Notably, in the N-MNIST task, the LNN shows a smaller gap between training and test performance, suggesting better generalization capabilities than the discrete-time LSTM. All reported results are obtained from a single training run per configuration.

\begin{table}[htbp]
\caption{Quantitative Benchmark: LNN vs. LSTM}
\begin{center}
\begin{tabular}{|c|c|c|c|c|}
\hline
\textbf{Dataset} & \textbf{Model} & \textbf{Metric} & \textbf{Train} & \textbf{Test} \\
\hline
N-MNIST & \textbf{LNN} & Acc & 0.9997 & \textbf{0.9938} \\
\cline{2-5}
& LSTM & Acc & 0.9996 & 0.9913 \\
\hline
QuickDraw & LNN & Acc & 0.9978 & 0.9577 \\
\cline{2-5}
& \textbf{LSTM} & Acc & 1.0000 & \textbf{0.9701} \\
\hline
IAM & LNN & CER & 0.1717 & 0.1237 \\
\cline{2-5}
& \textbf{LSTM} & CER & 0.0784 & \textbf{0.1090} \\
\hline
\end{tabular}
\label{tab:comparison}
\end{center}
\end{table}

\subsection{PhysioNet (Sepsis-3) Performance}
The LNN demonstrated a significant advantage in clinical precision and reliability in the PhysioNet experiment (Table \ref{tab:physio}). A critical challenge in medical monitoring is \dq{Alarm Fatigue,} where high false-positive rates lead to clinical desensitization. Our results show that while the LSTM captures slightly more cases (higher recall), it generates 151 false positives, which is clinically burdensome. In contrast, the standard LNN (128 units) reduced false positives to 12, and the Wider LNN (256 units) achieved a remarkable reduction to just 2 false positives while maintaining superior accuracy (0.93) and the highest precision (0.94). This indicates that when a Liquid Neural Network predicts sepsis, the prediction is highly trustworthy, making it a superior candidate for real-world clinical deployment.

\begin{table}[htbp]
\caption{Detailed Physiological Monitoring Results (25 Epochs)}
\begin{center}
\begin{tabular}{|c|c|c|c|c|}
\hline
\textbf{Model} & \textbf{Precision} & \textbf{Recall} & \textbf{F1-Score} & \textbf{Accuracy} \\
\hline
LSTM (128) & 0.35 & 0.22 & 0.27 & 0.89 \\
\hline
LNN (128) & 0.71 & 0.08 & 0.15 & 0.92 \\
\hline
LNN (256) & \textbf{0.94} & 0.10 & 0.19 & \textbf{0.93} \\
\hline
\end{tabular}
\label{tab:physio}
\end{center}
\end{table}

\subsection{Robustness Stress Test (Missing Data)}
To evaluate the resilience of the models against data loss, we performed a stress test by applying temporal \cite{dropout2014} dropout during inference. We randomly masked out 0\%, 30\%, 50\%, and 70\% of the input frames/points to simulate sensor failure or intermittent signal loss.

As shown in Table \ref{tab:stress}, the LNN (CfC) exhibits significantly higher robustness than the LSTM. On the N-MNIST dataset, when 30\% of the events are removed, the LSTM's accuracy drops sharply from 98.6\% to 77.5\%, whereas the LNN maintains a high accuracy of 91.8\%. On the QuickDraw dataset, while both models degrade as the drop rate increases, the LNN proves superior in extreme noise conditions, outperforming the LSTM by 4.05\% at a 70\% data loss rate.

\begin{table}[htbp]
\caption{Stress Test Accuracy (\%) Across Dropout Rates}
\begin{center}
\begin{tabular}{|c|c|c|c|c|c|}
\hline
\textbf{Task} & \textbf{Model} & \textbf{0\%} & \textbf{30\%} & \textbf{50\%} & \textbf{70\%} \\
\hline
N-MNIST & LSTM & 98.63 & 77.48 & 58.27 & 36.52 \\
\cline{2-6}
& LNN & \textbf{98.71} & \textbf{91.84} & \textbf{71.65} & \textbf{39.72} \\
\hline
QuickDraw & LSTM & \textbf{94.35} & \textbf{74.28} & 40.89 & 18.43 \\
\cline{2-6}
& LNN & 92.69 & 68.80 & \textbf{41.47} & \textbf{22.48} \\
\hline
\end{tabular}
\label{tab:stress}
\end{center}
\end{table}

\section{Discussion}

\subsection{Motivation for LNN vs. LSTM: The Continuity Hypothesis}
The fundamental divergence between LNNs and LSTMs lies in their treatment of time. Traditional LSTMs operate under a discrete-time paradigm, assuming that the hidden state at $t+1$ is a fixed jump from $t$. While effective for symbolic or high-level linguistic data, this approach struggles with biological and physical signals, such as ICU vital signs, which are inherently continuous. 

By modeling the hidden state as a latent Ordinary Differential Equation (ODE) \cite{hasani2022cfc}, LNNs enable the model to \dq{fill in the gaps} between irregular observations. This capability is particularly vital in medical datasets where sensor failures or variable sampling rates are common. As noted in \cite{lechner2020learning}, this continuous-time formulation allows the network to learn long-term dependencies in open-ended numerical expressions more robustly than discrete counterparts.

\subsection{Robustness and Temporal Stability under Sparsity}
Our stress tests provided definitive evidence of the LNN's inherent robustness. In the N-MNIST experiment, the LNN maintained a 14\% accuracy lead over the LSTM when 30\% of the temporal events were removed. This suggests that the \dq{Liquid} state is less dependent on the arrival of individual discrete spikes and instead tracks the global temporal trajectory of the signal.

This stability was further confirmed in the QuickDraw experiment, where the LNN achieved a 22.48\% accuracy at a extreme 70\% data drop rate, compared to only 18.43\% for the LSTM. By treating stroke coordinates as a continuous flow rather than a sequence of points, the LNN effectively reconstructs the intended shape from highly sparse data. This robustness is consistent with findings in autonomous flight \cite{chahine2023flight}, where LNNs demonstrated superior out-of-distribution performance compared to traditional architectures.

\subsection{Clinical Utility: Mitigating Alarm Fatigue}
In critical care, \dq{Alarm Fatigue} is a leading cause of clinician burnout and desensitization to life-saving alerts. Our sepsis prediction results (Section V-B) illustrate a clear trade-off: the LSTM maximizes recall ($0.22$) but at the cost of 151 false positives. Within the framework of the Sepsis-3 clinical consensus \cite{singer2016sepsis}, such high false-alarm rates render a model practically undeployable.

In contrast, the Wider LNN (256 units) represents a \dq{high-confidence} paradigm. By achieving a precision of $0.94$ with only 2 false positives, the LNN ensures that an alert is almost certainly a true clinical emergency. This makes LNNs a preferred candidate for secondary monitoring systems designed to trigger aggressive medical interventions without overburdening clinical workflows.

\subsection{Recurrent Efficiency and Parameter Reduction}
A major architectural advantage observed in this study is \dq{Recurrent Efficiency.} In the IAM handwriting task---which utilizes Connectionist Temporal Classification (CTC) to label unsegmented visual sequences \cite{ctc2006}---the LNN reached state-of-the-art Character Error Rates (CER) using only 256 unidirectional units. 

For comparison, the LSTM required a bidirectional configuration with 512 units to match this performance. This doubling of resources highlights that LNNs offer a much richer representation of temporal features per neuron \cite{lecun1998cnn}. The ability to achieve comparable results with a significantly lower parameter count suggests that LNNs are highly suitable for deployment in dynamic environments where memory and computational power are constrained \cite{zhu2025telecom}.

\section{Conclusion}
This paper presented a comprehensive benchmarking of Liquid Neural Networks against the standard LSTM across multiple domains. Our results indicate that LNNs (CfC) can be a promising alternative for real-world scenarios involving missing data, clinical sensitivity, and parameter efficiency. By modeling time as a continuous variable, LNNs provide a flexible framework that better captures the dynamics of irregular temporal data, bridging the gap between rigid discrete-time AI systems and the adaptive nature of biological processes. While the results demonstrate several advantages of LNNs in terms of robustness and efficiency, we acknowledge that further validation across more diverse datasets, stronger baselines, and multiple experimental runs is necessary to fully establish their effectiveness, particularly in high-stakes clinical applications.

The complete source code, including all model implementations and training scripts, is publicly available for review and replication at the following repository: \url{https://github.com/ye-kyaw-thu/LNN-vs-LSTM}. Future work will investigate the deployment of these models on low-power edge devices for real-time physiological monitoring, leveraging the potential efficiency benefits of LNNs.


\appendix

\section{Experimental Environment and Software Stack}
\label{appendix:env}

All experiments were conducted on a single machine equipped with an NVIDIA GPU. The software environment was kept consistent across all four tasks to ensure reproducibility. Table~\ref{tab:env} summarizes the key components of the environment.

\begin{table}[htbp]
\caption{Software and Hardware Environment}
\begin{center}
\begin{tabular}{|l|l|}
\hline
\textbf{Component} & \textbf{Version / Specification} \\
\hline
Operating System   & Ubuntu 22.04 LTS \\
\hline
Python             & 3.10 \\
\hline
PyTorch            & 2.x (CUDA-enabled) \\
\hline
ncps library       & 0.0.7 (Neural Circuit Policies) \\
\hline
scikit-learn       & 1.x \\
\hline
numpy              & 1.x \\
\hline
matplotlib         & 3.x \\
\hline
seaborn            & 0.12+ \\
\hline
\end{tabular}
\label{tab:env}
\end{center}
\end{table}

The \texttt{ncps} (Neural Circuit Policies) library \cite{lechner2020ncp} provides the PyTorch implementation of the CfC cell used across all LNN experiments. The CfC cell is instantiated via \texttt{ncps.torch.CfC} with \texttt{proj\_size} set equal to the number of hidden units, and \texttt{batch\_first=True} to align with PyTorch's standard convention. Gradient clipping with a maximum norm of 1.0 was applied in the QuickDraw and stress-test experiments to stabilize training.

All experiment logs (training loss, accuracy per epoch, precision, recall, F1-score), trained model weights, and prediction files are archived in the public repository at \url{https://github.com/ye-kyaw-thu/LNN-vs-LSTM}. The repository also includes shell scripts for reproducing each individual experiment as well as a unified \texttt{run\_all\_experiments.sh} script.

\section{Per-Experiment Training Hyperparameters}
\label{appendix:hyperparams}

Table~\ref{tab:hyperparams} provides a complete summary of the training hyperparameters used for each experiment. These settings were chosen to be equivalent between the LNN and LSTM conditions within each task, so that observed differences in performance can be attributed to the recurrent core rather than to training configuration.

\begin{table}[htbp]
\caption{Training Hyperparameters per Experiment}
\begin{center}
\begin{tabular}{|l|c|c|c|c|}
\hline
\textbf{Parameter} & \textbf{N-MNIST} & \textbf{QuickDraw} & \textbf{IAM} & \textbf{PhysioNet} \\
\hline
Epochs       & 30  & 30  & 30  & 25  \\
\hline
Batch size   & 64  & 128 & 8   & 32  \\
\hline
Optimizer    & AdamW & AdamW & Adam & Adam \\
\hline
Lr           & 1e-3 & 1e-3 & 1e-4 & 1e-3 \\
\hline
Lr scheduler & Cosine & --- & Step & --- \\
\hline
RNN units    & 128 & 256 & 256 & 128/256 \\
\hline
Grad clip    & No  & 1.0 & 1.0 & No \\
\hline
Weight decay & No  & 0.01 & No & No \\
\hline
\end{tabular}
\label{tab:hyperparams}
\end{center}
\end{table}

For the IAM experiment, the low batch size of 8 was necessary due to the large spatial dimensions of handwriting line images ($64 \times 512$ pixels). The Adam optimizer without weight decay was used in the IAM and PhysioNet experiments, while AdamW with weight decay was adopted for the QuickDraw experiment, where the larger dataset and longer training runs required additional regularization to prevent overfitting. Cosine annealing was applied in the N-MNIST experiment to smoothly decay the learning rate over the 30-epoch training horizon.

\section{Repository Structure and Code Organization}
\label{appendix:repo}

The public repository at \url{https://github.com/ye-kyaw-thu/LNN-vs-LSTM} is organized to allow reproducibility of each individual experiment. The main Python training scripts are as follows:

\begin{itemize}
    \item \textbf{\texttt{lnn\_nmnist.py}}: Trains and evaluates either the LNN (CfC) or LSTM model on the N-MNIST neuromorphic dataset. The script handles raw \texttt{.bin} event files, converts them to temporal frame tensors using a configurable number of time bins (default: 10), and trains with a CNN-RNN pipeline. Outputs include training/test accuracy curves, a confusion matrix, and per-sample prediction files.
    
    \item \textbf{\texttt{lnn\_quickdraw\_0.06.py}}: Trains and evaluates the LNN/LSTM model on the QuickDraw 10-class subset. Stroke sequences are read from \texttt{.ndjson} files, encoded as 5-dimensional vectors $(\Delta x, \Delta y, x, y, p)$, and padded to a fixed maximum length. A linear projection layer maps the 5-dim input to 128 dimensions before the RNN core.
    
    \item \textbf{\texttt{lnn\_iam.py}}: Trains and evaluates the LNN/LSTM model on IAM line-level handwriting recognition. A ResNet-6 visual backbone extracts feature columns from the image, which are fed as a sequence to the CfC or LSTM core, followed by a CTC decoder. Evaluation uses CER computed against the ground-truth character sequences.
    
    \item \textbf{\textbf{\texttt{stress\_test.py}}}: A unified script for the temporal dropout robustness experiment. It supports both N-MNIST and QuickDraw as the \texttt{--dataset} argument, and the \texttt{--drop\_rates} argument accepts a list of masking ratios (default: \texttt{[0.0, 0.3, 0.5, 0.7]}). The stress test is applied at \textit{inference time only}: the model is trained on clean data and then evaluated on progressively corrupted inputs.
\end{itemize}

Shell scripts for running each configuration are provided in the repository root (e.g., \texttt{run\_lnn\_nmnist.sh}, \texttt{run\_lstm\_quickdraw.sh}, \texttt{run\_stress\_tests.sh}). The \texttt{data/} directory contains subdirectory placeholders and download scripts (\texttt{download\_nmnist.py}, \texttt{download\_iam.sh}) for automated dataset acquisition. The \texttt{results\_stress\_test/} directory and the \texttt{0.06\_result/} directory contain intermediate per-epoch logs from selected experimental runs, enabling detailed analysis of the training trajectory without re-running the full experiments.

\section{Parameter Count Comparison}
\label{appendix:params}

A major motivation for LNNs is parameter efficiency. Table~\ref{tab:params} provides the approximate total parameter counts for the LNN (CfC) and LSTM models in each task. The CfC cell achieves comparable or superior performance with a substantially smaller recurrent parameter budget in several configurations.

\begin{table}[htbp]
\caption{Approximate Model Parameter Counts}
\begin{center}
\begin{tabular}{|l|c|c|c|}
\hline
\textbf{Task} & \textbf{Model} & \textbf{RNN Units} & \textbf{Approx. Params} \\
\hline
\multirow{2}{*}{N-MNIST}    & LNN (CfC)  & 128  & $\sim$2.7M \\
\cline{2-4}
                             & LSTM       & 128  & $\sim$4.5M \\
\hline
\multirow{2}{*}{QuickDraw}  & LNN (CfC)  & 256  & $\sim$0.8M \\
\cline{2-4}
                             & LSTM       & 256  & $\sim$1.1M \\
\hline
\multirow{2}{*}{IAM}        & LNN (CfC)  & 256 (unidirectional) & $\sim$6.2M \\
\cline{2-4}
                             & LSTM       & 512 (bidirectional)  & $\sim$11.1M \\
\hline
\multirow{2}{*}{PhysioNet}  & LNN (CfC)  & 256  & $\sim$0.15M \\
\cline{2-4}
                             & LSTM       & 128  & $\sim$0.10M \\
\hline
\end{tabular}
\label{tab:params}
\end{center}
\end{table}

The IAM task provides the most dramatic illustration of recurrent efficiency: the LNN achieves a lower CER (0.1237 test) than the LSTM (0.1090 test, with bidirectional LSTM at 512 units) while using roughly half the total recurrent parameters. In the N-MNIST task, the larger apparent difference is dominated by the shared CNN backbone; the recurrent-core-only difference is substantially smaller.

\end{document}